# A survey on joint object detection and pose estimation using monocular vision

**Aniruddha V Patil[1] and Pankaj Rabha[2]**

[1] IIIT-Hyderabad, Gachibowli, Hyderabad, Telangana, India 500032
[2] Intel, Bellandur, Bangalore, Karnataka, India 560103

aniruddha.patil@students.iiit.ac.in, pankaj.rabha@intel.com

**Abstract**. In this survey we present a complete landscape of joint object detection and pose estimation methods that use monocular vision. Descriptions of traditional approaches that involve descriptors or models and various estimation methods have been provided. These descriptors or models include chordiograms, shape-aware deformable parts model, bag of boundaries, distance transform templates, natural 3D markers and facet features whereas the estimation methods include iterative clustering estimation, probabilistic networks and iterative genetic matching. Hybrid approaches that use handcrafted feature extraction followed by estimation by deep learning methods have been outlined. We have investigated and compared, wherever possible, pure deep learning based approaches (single stage and multi stage) for this problem. Comprehensive details of the various accuracy measures and metrics have been illustrated. For the purpose of giving a clear overview, the characteristics of relevant datasets are discussed. The trends that prevailed from the infancy of this problem until now have also been highlighted.

## 1. Introduction

Object detection is the process of finding instances of real-world objects in images or videos. Object detection algorithms typically use extracted features and learning algorithms to recognize instances of an object category.

The position and orientation of an object is called the pose of the object. To determine the pose of an object in an image is called pose estimation. A general task in the various applications of pose estimation is to determine the position and orientation of each of the object in the scene with regard to some coordinate system. There have been many works on object detection and pose estimation separately. This work aims to cover all works that treat these two problems simultaneously.

Monocular vision, by definition is vision with one eye/camera. This type of vision greatly limits the perception of depth. For all of joint object detection and pose estimation, the goal remains the same, given an input image, provide what objects exist in the input image and their pose. The methods used vary based on the level of sophistication of the input. In this paper, we discuss methods where the input is a single image (RGB or Grayscale).

The various applications of this problem include augmented reality [1], autonomous driving [2], robotic manipulation [3,4] and inventory management [5]. Applications like augmented reality and autonomous driving need real-time approaches whereas this constraint can be relaxed a little when it comes to robotic manipulation or inventory management in situations where taking time does not pose a threat.

## 2. Input types
Advancement in technology has enabled us to capture input data at various sophistication levels.

### 2.1. Monocular (Colored/Grayscale)
One can say that this is the most common form of input. A colored monocular image is easily available from any camera. Medical applications that interpret signals as images commonly use grayscale images.

### 2.2. RGB-D
These are colored images along with depth data. The apparatus required for these images has recently become more economically viable in the form of the Microsoft Xbox Kinect. Other popular sensors are the Asus Xtion PRO LIVE and The Leap.

### 2.3. LiDAR
Light Detection and Ranging is an optical remote-sensing technique that uses laser light to densely sample the surroundings, producing accurate measurements of distance to each point the laser hits. The major hardware components of the LiDAR system include a vehicle, laser scanner system, Global Positioning System (GPS) and an Inertial Navigation System (INS).

### 2.4. RADAR
Radio Detection and Ranging is a radio wave based detection system that helps determine the location and velocity of objects. There exist three kinds of RADAR - Short Range (SRR), Medium Range (MRR) and Long Range (LRR). In the context of autonomous driving, these RADAR systems perform different tasks (SRR for object/pedestrian detection and emergency braking, MRR for lane change assistance and cross traffic alert and LRR for Autonomous Cruise Control (ACC) and collision warning).

## 3. Formulation of object detection and pose estimation

### 3.1. Object detection
Object detection has two main components – localization and classification of the object of interest. Localization can be formulated as estimating a vector $(x, y, h, w)$, where $x, y, h$ and $w$ are respectively the x-coordinate, y-coordinate, height and width of the object. These are mostly specified in the camera coordinate system. Classification can be formulated as estimating a single number $c$ that corresponds to the class number of the object of interest. Combining these two, object detection can be formulated as estimating a vector $(c, x, y, h, w)$.

### 3.2. Pose estimation
There are two ways in which we can formulate the pose estimation problem.
We can formulate it as a classification problem where the all the possible configurations of the object are quantized into various classes. The pose of the object is then classified into the appropriate class.

Assume that the set of all possible configurations of the object is $D$ then, on dividing this set into $k$ dimensions such that all possible configurations in the $i^{th}$ dimension can be depicted as $D^i$ where $i \in \{1..k\}$. Further division of $D^i$ into $n$ subsets gives us $D^i_j$ where $j \in \{1..n\}$. Thus, the problem becomes to correctly place the pose of the object into one of the $D^i_j$ classes.

In the formulation of pose estimation as a regression problem, we regress an initial pose to best fit the pose of the object. When we assume the pose of the object to be $k$-dimensional, the regression-based approach should estimate a $k$-dimensional vector as the pose of the object. Now we look at different types of approaches for joint object detection and pose estimation. They can be categorized into vision based, hybrid and deep learning based approaches.

## 4. Vision based

The vision based approaches heavily use domain knowledge and involve modeling the input space in such a way that the expected output can be learnt based on the model. The mathematical formulation, information contained in the selected features, determining what features to use and using the right learning technique for efficiency are key to approach this problem with this method.

The different contributions of works that fall under this approach can be broadly categorized into descriptors/models and estimation methods.

### 4.1. Descriptors/Models

#### 4.1.1. Chordiograms

Are shape representations based on geometric relationships of object boundary edges ensuring that foreground elements are prioritized. A chordiogram is a $k$ dimensional histogram of all chord features on the boundary of a segmented object. A chord is a pair of points $(p,q)$ on the boundary points. The chord feature $d_{pq}$ is given as

$$d_{pq} = (l_{pq}, \varphi_{pq}, \theta_p - \varphi_{pq}, \theta_q - \varphi_{pq}) \qquad (1)$$

Where $l_{pq}$ is the chord length, $\varphi_{pq}$ is the chord orientation and $\theta_p$ and $\theta_q$ are normals of the object boundary at $p$ and $q$. They take into account the chord vector lengths, their orientations and normals given the silhouette of the object of interest and capture the geometric relationships like relative location and normals between pairs of boundary edges as a high-level descriptor [6].

#### 4.1.2. Shape aware deformable parts model

A deformable parts model (DPM) is a star shaped conditional random field (CRF), with a root part for the whole object and several parts connected to the root. Each model part has an anchor point which is relative to the root part. These parts are allowed to move around their anchor point with a penalty. For simultaneous object detection and pose estimation from only model silhouette information, a shape aware modification of the DPM can be made. Every coarse pose of the object maps to a component of the learned S-DPM. By grouping nearby viewpoints, a certain number of discrete poses are arrived at. Silhouettes of a coarse pose cluster are used in the training of a S-DPM component as positives. The silhouettes of other poses, objects and random background edges are used as negatives. The coarse pose of the object is determined by the component with the maximum score at each image location [6].

#### 4.1.3. Bag of boundaries (BOB)

After extracting contours of the image containing the object of interest, the bounding boxes of various views of the object are resized to their average size. A template is created in which each pixel counts the number of contours it is a part of. This gives us a midlevel representation of the image as a bag of boundaries for shape matching [7]. A BOB operates on information apart from shape context. It operates on image edge information by detecting and aggregating boundaries in the neighbourhood of the BOB. The best matching subset of BOBs to the given shape templates, such that maximal coverage of the image contours estimated to be boundaries, are found in the case of object detection.

#### 4.1.4. Distance transform templates

Templates based on the distance transform have proved to be effective in detecting and estimating the pose of texture less planar objects. In [8], an edge map is created with the Canny edge detector [9], followed by applying the distance transform on this edge map. The distance transform template is then created by computing the level curve at a given value. The level curve is essential as it makes sure that small gaps do not cause the object of interest to be detected as two distinct objects.

*4.1.5. Natural 3D markers*
Natural 3D markers [10] are a set of landmarks on the object of interest. They are chosen based on their visibility from different views and that are robust to illumination, partial occlusions, clutter, etc. They are extracted under multiple viewpoints. In the case of the N3Ms being clustered in a single region, the occlusion of that region would cause this method to fail. Thus, the points are taken such that they are evenly spread across the entire object of interest.

*4.1.6. Facet features*
Given the image contours, polygons are approximated to reduce the number of vertices. Prior knowledge of the facets of the object of interest is then used for detection and pose estimation [1].

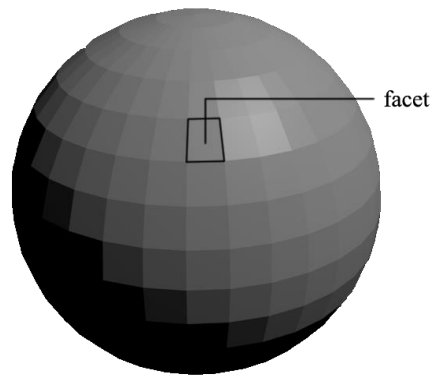

**Figure 1.** Depiction of a facet.

*4.2. Estimation methods*

*4.2.1. Expectation Maximization (EM) and Iterative Clustering Estimation (ICE)*
EM finds the maximum likelihood parameters in statistical models. These methods are conceptually similar (ICE is the manifestation of EM that considers local features that map to an object instance and evaluates pose hypotheses). The expected membership of each data point to one of the distributions is given by the expectation step and the parameters for each distribution are given by the maximization step. Similarly, the expected membership of each local feature to an object instance is given by the clustering step and the best object pose corresponding to the feature memberships is given by the estimation step [4,7].

*4.2.2. Probabilistic Network*
This tree-structured network of probabilistic classifiers uses boosting for quick rejection of negatives while the positives traverse multiple branches. The fundamental total probability law is leveraged to compute the probability of being an object, given the pose parameters [11].

*4.2.3. Iterative genetic matching*
This method uses a genetic algorithm with a given population, crossover rate and mutation rate for the purpose of matching. The matching rate is the metric used for determining the fitness of an individual. This method is proposed as an alternative to the EM and ICE methods [12].

## 5. Hybrid approaches

In the case of hybrid approaches, the handcrafted features, representations or intermediate results of some other technique work with a neural network. Again, the features are as important as the architecture of the network to make this approach accurate and efficient.

Geometric constraints applied to a predicted 2D bounding box followed by regression by a CNN to give a 3D bounding box is shown in [2]. Though hybrid methods try to achieve best of both worlds, extending the network to also extract features has proved to be much more effective.

## 6. Pure deep learning based approaches

Pure deep learning based approaches use end-to-end networks to map the inputs directly to the expected outputs without any intermediate representations or extracted features. Though computationally expensive, due to recent breakthroughs in technology, accurate real-time approaches are feasible on high-end GPUs. Recently, CNNs have been outperforming DPM based methods for the task at hand. We can further classify these approaches into single stage or multistage methods.

A network is said to be multistage if it has stages performing distinct tasks linearly. A single stage network has no intermediates and provides the output directly from the input.

### 6.1. Single stage

SSD-6D [13] proposes an extension of the SSD [14] detection framework for the task of 3D detection and 3D rotation estimation. The 3D rotation space is decomposed into discrete viewpoints and in-plane rotations. The rotation estimation is then treated as a classification problem. Appropriate sampling for the rotation space is required for obtaining good results. This approach requires an offline stage to estimate 3D translation. This stage is used to precompute bounding boxes w.r.t. all possible sample rotations.

PoseCNN [15] jointly carries out segmentation and pose estimation of the segmented objects. It relies on a fully convolutional network to localize objects which might make it harder to deal with multiple instances of the same object.

Deep-6DPose [16] proposes adding a branch to the prominent Mask-RCNN [17] for the purpose of pose estimation such that no further pose refinement is necessary. Decoupling of pose parameters into translation and rotation so that the rotation can be regressed via a Lie algebra representation has been shown.

In [18], the authors attempt to combine detection and pose estimation at the same level. The scores for the presence of an object category, the offset for its location and the approximate pose are all estimated on a regular grid of locations in the image.

### 6.2. Multi stage

In BB8 [19], the idea of a cascade of multiple CNNs is proposed. In the first stage, a segmentation network is applied to the input image to localize objects. The next stage involves another CNN to predict 2D projections of the corners of the 3D bounding boxes around objects. Pose in 6D is then estimated for the correspondences between the projected 2D coordinates and the 3D ground control points of bounding box corners.

## 7. Accuracy measures

The accuracy measures are specific to the proposed methods in the earlier works of pose estimation. Intersection over Union (IoU) and Mean Average Precision (mAP) has been the standard for reporting object detection accuracies. For pose estimation as a classification problem, mAVP is used, whereas for regression, 2D Projections [21], 5cm5° [22] and ADD (or 6D pose) [23] are used.

### 7.1. Object detection

#### 7.1.1. IoU

IoU gives a measure of overlap between the detected and the ground truth bounding boxes. It has become the standard for reporting detection accuracies. Generally, 0.5 and 0.9 are the thresholds used for considering a detection as successful.

*7.1.2. mAP*

Average precision (AP) is the average of maximum precision values across eleven recall values (from 0 to 1.0) with a step of 0.1. Then the mean average precision is the mean of APs across all classes.

$$AP = \frac{1}{11}(AP_r(0) + AP_r(0.1) + AP_r(0.2) + \cdots + AP_r(1.0)) \qquad (2)$$

*7.2. Pose estimation*

*7.2.1. mAVP*

AVP is an extension of the AP metric that includes the extra constraint that the pose classification is also correct, in addition to the IoU being greater than 0.5. AVP is evaluated at different levels according to how many total pose bins exist. It is generally evaluated at 4, 8, 12 and 24 bins. The average AVP across all the objects is the mAVP of the pipeline.

*7.2.2. 2D projections*

A pose is considered correct if the average of the 2D distances between the projections of the object's vertices from the estimated pose and the ground truth pose is less than 5 pixels.

*7.2.3. 5cm5°*

The 5cm5° metric states that the predicted pose is said to be correct if it is within 5cm of translational error and 5° of rotational error of the ground truth.

*7.2.4. ADD or 6D pose*

The ADD metric states that an estimated pose is accepted if the average distance between the ground truth pose and the estimated pose is smaller than 10% of the diameter of the object.

$$ADD = \frac{1}{m} \sum_{x \in \mathcal{M}} \|(Rx + T) - (R\tilde{x} + \tilde{T})\| \qquad (3)$$

Where $T$, $R$, $\tilde{T}$ and $\tilde{R}$ are the respective ground truth translation, ground truth rotation, estimated translation and estimated rotation of the 3D model $\mathcal{M}$.

## 8. Datasets

Many approaches have proposed their own datasets for this task, however over time, LineMOD [23] and the Tejani *et al.* dataset [24] have become popular. The T-LESS dataset [25] has come up recently with a large number of instances per class using various imaging tools. More specific datasets like OCCLUSION [26] are trained on for making the pipelines robust to occlusion.

*8.1. LineMOD*

It has over 18,000 images with 15 different objects and labels for ground truth pose. It also has the 3D models saved as a point clouds.

*8.2. Tejani et al.*

Has around 5,000 images with 6 object categories. There are an average of 870 instances per category of indoor setting. It also has non-centred objects and cluttered backgrounds. Labels are provided for orientation.

*8.3. Pascal3D+*

The Pascal3D+ dataset [20] has 12 object categories with around 3,000 instances per category of both indoor and outdoor setting. It also has non-centred objects and backgrounds that are cluttered. Labels are provided for occlusion level and orientation.

*8.4. T-LESS*

This dataset has around 38,000 training and 10,000 test images from each sensor of 30 different textureless objects. The training images are of the objects against a black background and the testing images are real scenes ranging from simple to the ones with high clutter and occlusion. The sensors used are Primesense CARMINE 1.09 (a structured-light RGB-D sensor), Microsoft Kinect v2 (a time-of-flight RGB-D sensor), and Canon IXUS 950 IS (a high-resolution RGB camera). They provide a manually created CAD model and a semi-automatically reconstructed one.

**Table 1.** Characteristics of the discussed datasets

| Dataset | Images | Classes | Instances | Scene Type |
|---|---|---|---|---|
| LineMOD [23] | ~18,000 | 15 | ~1,200 | Indoor |
| Tejani et al. [24] | ~5,000 | 6 | ~870 | Indoor |
| Pascal3D+ [20] | ~36,000 | 12 | ~3,000 | Indoor and outdoor |
| T-LESS [25] | ~48,000 | 30 | ~1,600 | Indoor |

**Table 2.** The reported results of the discussed methods on the discussed datasets

| Dataset | Method | Measure | Value of Measure |
|---|---|---|---|
| LineMOD [23] | SSD-6D [13] | IoU (0.5) | 99.4 |
| | | ADD | 76.3 |
| | Deep6DPose [16] | IoU (0.5) | 99.7 |
| | | 2D projection | 99.3 |
| | | 5cm5° | 68.5 |
| | BB8 [19] | 2D projection | 89.3 |
| | | ADD | 62.7 |
| | | 5cm5° | 69 |
| Tejani et al. [24] | Deep6DPose [16] | IoU (0.5) | 99.6 |
| | | 2D projection | 99.3 |
| | | ADD | 62 |
| | | 5cm5° | 64.5 |
| | SSD-6D [13] | IoU (0.5) | 96.3 |
| | | 2D projection | 98.8 |
| Pascal3D+ [20] | Share300 [18] | mAVP (24) | 27.7 |
| T-LESS [25] | BB8 [19] | ADD | 54 |

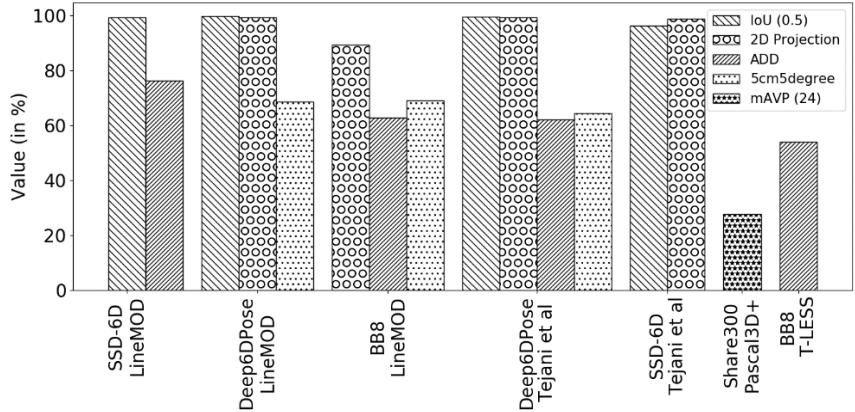

**Figure 2.** A graphical depiction of Table 2

## 9. Trends

*9.1. Accumulation of data*
There was a scarcity of data when the relatively old works were published, hence they all proposed their own dataset and accuracy measures with respect to datasets in their experiments. Hence, there was no normalized method to report accuracies. Over time, the works have consistently performed experiments on the LineMOD and Tejani *et al.* [24] datasets. The accuracy measures of 5cm5° and ADD have become the standards for pose estimation, whereas IoU and mAP are used for detection.

*9.2. Evolution of approaches*
The earlier works showcased their accuracy and the feasibility of their representations/methods using metrics specific to the dataset used. On the other hand, in the recent times, feasibility has become a bare minimum, in addition to which, methods must prove that their accuracies and speeds are also feasible using standard metrics on standard datasets compared to existing methods.

## 10. Problems

*10.1. Occlusion*
An object is said to be occluded if its contour cannot be seen completely as it is obstructed by another object. When the properties of models are learnt or modelled considering the object as a whole, there may be severe drops in their accuracies due to occlusion. To overcome this, DPMs have come up that allow predictions based on the visible parts of the object. As mentioned earlier, since CNNs implicitly model the object parts unlike the explicit modelling as in the case of DPMs, CNNs tend to perform better than DPM based models for occluded objects also. The use of higher potentials for detection and pose estimation along with segmentation is described in [21]. It is based on HOG templates for objects and occlusion that do not model a specific object that causes occlusion. Hence they are termed as discriminatively learnt HOG templates.

*10.2. Illumination*
Varying types of illumination that cause varying pixel intensities may cause issues with the extracted features and learnt models. One way to get over this problem is normalization during learning and inference. However, in the case that only the object in the scene is poorly illuminated, normalization will not help. One can borrow from the idea of [10] to use evenly distributed keypoints across the object that are invariant to illumination for the purpose of its detection and pose estimation.

*10.3. Clutter*

The objects of interest must be detected regardless of the background. Learning the properties of the object against a fixed background has been shown to cause a drop in performance during inference. Thus, the randomness that clutter introduces in the background is necessary for robustness. Ideally, the training data should have different levels of clutter from zero to heavy for good accuracies at all the possible levels of clutter. Methods proposed in [15] are robust to cluttered scenes.

*10.4. Scene diversity*

Scene diversity has quite an impact on the performance. If the learning process is carried out only in a fixed environment, there is less variation in the scene properties, which causes the inference in another environment to suffer in terms of accuracy. Training must be carried out in all the possible environments in which inference would be carried out. The use of PASCAL3D+ [20] that has both indoor and outdoor environments is shown in [18] to make the system more robust to scene diversity.

## 11. Conclusion

Through this survey, we have provided an overview of the developments in joint object detection and pose estimation approaches that use monocular vision. We outlined descriptors/models and estimation methods used in traditional approaches. A brief illustration of hybrid approaches was made. Single stage and multi stage pure deep learning based pipelines were discussed. Characteristics of various datasets were discussed and the standard accuracy measures were illustrated. Wherever possible, statistics of the datasets and results on them were tabulated. The trends from the infancy of this problem up until now were also explained.

**References**
[1] Shahrokni A, Vacchetti L, Lepetit V and Fua P 2002 Polyhedral object detection and pose estimation for augmented reality applications *Proc. of Computer Animation* (Lausanne: Geneva) pp 65–9
[2] Mousavian A, Anguelov D, Flynn J and Košecká J 2017 3d bounding box estimation using deep learning and geometry *IEEE Conf. on Computer Vision and Pattern Recognition* (Fairfax: Honolulu) pp 5632–40
[3] Collet A, Berenson D, Srinivasa SS and Ferguson D 2009 Object recognition and full pose registration from a single image for robotic manipulation *IEEE Int. Conf. on Robotics and Automation* (Pittsburgh: Kobe) pp 48–55
[4] Collet A, Martinez M and Srinivasa SS 2010 The moped framework: object recognition and pose estimation for manipulation *IEEE Int. Conf. on Robotics and Automation* (Pittsburgh: Anchorage) pp 1284–306
[5] Lim JJ, Pirsiavash H and Torralba A 2013 Parsing ikea objects: fine pose estimation *IEEE Int. Conf. on Computer Vision* (Cambridge: Sydney) pp 2992–9
[6] Zhu M, Derpanis KG, Yang Y, Brahmbhatt S, Zhang M, Phillips C, Lecce M and Daniilidis K 2014 Single image 3d object detection and pose estimation for grasping *IEEE Int. Conf. on Robotics and Automation* (Philadelphia: Hong Kong) pp 3936–43
[7] Payet N and Todorovic S 2011 From contours to 3d object detection and pose estimation *IEEE Int. Conf. on Computer Vision* (Corvallis: Barcelona) pp 983–90
[8] Holzer S, Hinterstoisser S, Ilic S and Navab N 2009 Distance transform templates for object detection and pose estimation *IEEE Conf. on Computer Vision and Pattern Recognition* (Garching: Miami) pp 1177–84
[9] Canny J 1986 A computational approach to edge detection *IEEE Trans. on Pattern Analysis and Machine Intelligence* **6** 679–98
[10] Hinterstoisser S, Benhimane S and Navab N 2007 N3m: natural 3d markers for real-time object detection and pose estimation (Garching: Rio de Janerio) *IEEE Int. Conf. on Computer Vision* pp 1–7
[11] Zhang J, Zhou SK, McMillan L and Comaniciu D 2007 Joint real-time object detection and

pose estimation using probabilistic boosting network *IEEE Conf. on Computer Vision and Pattern Recognition* (Chapel Hill: Minneapolis) pp 1–8
[12] Kayanuma M and Hagiwara M 1999 A new method to detect object and estimate the position and the orientation from an image using a 3-d model having feature points *IEEE Int. Conf. on Systems, Man, and Cybernetics* (Yokohama: Tokyo) pp 931–6
[13] Kehl W, Manhardt F, Tombari F, Ilic S and Navab N 2017 Ssd-6d: making rgb-based 3d detection and 6d pose estimation great again *IEEE Int. Conf. on Computer Vision* (Munich: Venice) pp 1530–8
[14] Liu W, Anguelov D, Erhan D, Szegedy C, Reed S, Fu CY and Berg AC 2016 Ssd: single shot multibox detector *European Conf. on Computer Vision* (Chapel Hill: Amsterdam) pp 21–37
[15] Xiang Y, Schmidt T, Narayanan V and Fox D 2017 arXiv:1711.00199
[16] Thanh-Toan D, Ming C, Trung P and Ian R, 2018 arXiv:1802.10367
[17] Kaiming H, Gerogia G, Piotr D and Ross G 2017 Mask r-cnn *IEEE Int. Conf. on Computer Vision* (Menlo Park: Venice) pp 2980-8
[18] Poirson P, Ammirato P, Fu CY, Liu W, Kosecka J and Berg AC 2016 Fast single shot detection and pose estimation *4th Int. Conf. on 3D Vision* (Chapel Hill: Stanford) pp 676–84
[19] Rad M and Lepetit V 2017 Bb8: a scalable, accurate, robust to partial occlusion method for predicting the 3d poses of challenging objects without using depth *IEEE Int. Conf. on Computer Vision* (Graz: Venice) pp 3848–56
[20] Xiang Y, Mottaghi R and Savarese S 2014 Beyond pascal: a benchmark for 3d object detection in the wild *IEEE Winter Conf. on Applications of Computer Vision* (Ann Arbor: Steamboat Springs) pp 75–82
[21] Brachmann E, Michel F, Krull A, Ying Yang M and Gumhold S 2016 Uncertainty-driven 6d pose estimation of objects and scenes from a single rgb image *IEEE Conf. on Computer Vision and Pattern Recognition* (Dresden: Las Vegas) pp 3364–72
[22] Shotton J, Glocker B, Zach C, Izadi S, Criminisi A and Fitzgibbon A 2013 Scene coordinate regression forests for camera relocalization in rgb-d images *IEEE Conf. on Computer Vision and Pattern Recognition* (Cambridge: Portland) pp 2930–7
[23] Hinterstoisser S, Lepetit V, Ilic S, Holzer S, Bradski G, Konolige K and Navab N 2012 Model based training, detection and pose estimation of texture-less 3d objects in heavily cluttered scenes *Asian Conf. on Computer Vision* (Munich: Daejeon) pp 548–62
[24] Tejani A, Tang D, Kouskouridas R and Kim TK 2014 Latent-class hough forests for 3d object detection and pose estimation *European Conf. on Computer Vision* (London: Zurich) pp 462–77
[25] Hodan T, Haluza P, Obdržálek Š, Matas J, Lourakis M and Zabulis X 2017 T-less: An rgb-d dataset for 6d pose estimation of texture-less objects *IEEE Winter Conf. on Applications of Computer Vision* (Prague: Santa Rosa) pp 880–8
[26] Brachmann E, Krull A, Michel F, Gumhold S, Shotton J and Rother C 2014 Learning 6d object pose estimation using 3d object coordinates *European Conf. on Computer Vision* (Dresden: Zurich) pp 536-551